\documentclass[conference]{IEEEtran}
\IEEEoverridecommandlockouts
\usepackage{cite}
\usepackage{amsmath,amssymb,amsfonts}
\usepackage{algorithmic}
\usepackage{graphicx}
\usepackage{textcomp}
\usepackage{listings}
\usepackage{adjustbox}
\usepackage{xcolor}
\def\BibTeX{{\rm B\kern-.05em{\sc i\kern-.025em b}\kern-.08em
    T\kern-.1667em\lower.7ex\hbox{E}\kern-.125emX}}
\usepackage{booktabs}

\definecolor{codegreen}{rgb}{0,0.6,0}
\definecolor{codegray}{rgb}{0.5,0.5,0.5}
\definecolor{codepurple}{rgb}{0.58,0,0.82}
\definecolor{backcolour}{rgb}{0.95,0.95,0.92}

\let\othelstnumber=\thelstnumber
\def\createlinenumber#1#2{
    \edef\thelstnumber{%
        \unexpanded{%
            \ifnum#1=\value{lstnumber}\relax
              #2%
            \else}%
        \expandafter\unexpanded\expandafter{\thelstnumber\othelstnumber\fi}%
    }
    \ifx\othelstnumber=\relax\else
      \let\othelstnumber\relax
    \fi
}

\lstdefinestyle{customc}{
  belowcaptionskip=1\baselineskip,
  breaklines=true,
  frame=single,
  xleftmargin=0.35cm,
  xrightmargin=0.15cm,
  numbers=left,
  numbersep=5pt,  
  language=C,
  showstringspaces=false,
  basicstyle=\footnotesize\ttfamily,
  keywordstyle=\bfseries\color{green!40!black},
  commentstyle=\itshape\color{purple!40!black},
  identifierstyle=\color{blue},
  stringstyle=\color{orange},
}

\lstdefinestyle{customcArianeExploit1}{
  breaklines=true,
  frame=single,
  xleftmargin=0.4cm,
  xrightmargin=0.2cm,
  numbers=left,
  numbersep=5pt,  
  language=C,
  showstringspaces=false,
  basicstyle=\footnotesize\ttfamily,
  keywordstyle=\bfseries\color{green!40!black},
  commentstyle=\itshape\color{purple!60!black},
  identifierstyle=\color{blue},
  stringstyle=\color{yellow!50!black},
  morekeywords={asm},
  keywordstyle=[2]\bfseries\color{brown!60!black},
}

\lstdefinestyle{customcArianeExploit}{
  breaklines=true,
  frame=single,
  xleftmargin=0.4cm,
  xrightmargin=0.2cm,
  numbers=left,
  numbersep=5pt,  
  language=C,
  showstringspaces=false,
  basicstyle=\footnotesize\ttfamily,
  keywordstyle=\bfseries\color{blue},
  commentstyle=\itshape\color{green!50!black},
  identifierstyle=\color{black},
  stringstyle=\color{brown},
  morekeywords={asm},
  keywordstyle=[2]\bfseries\color{black},
}

\lstdefinestyle{customlog}{
  breaklines=true,
  frame=single,
  xleftmargin=0.35cm,
  xrightmargin=0.15cm,
  numbers=left,
  numbersep=5pt,  
  language=C,
  showstringspaces=false,
  basicstyle=\footnotesize\ttfamily,
  keywordstyle=\color{blue},
  commentstyle=\itshape\color{purple!40!black},
  identifierstyle=\color{blue},
  stringstyle=\color{orange},
  keywords=[2]{INFO},
  keywords=[3]{ERROR},x
  keywordstyle=[2]\bfseries\color{green!40!black},
  keywordstyle=[3]\bfseries\color{red!500!black},
}

\definecolor{verilogcommentcolor}{RGB}{0,124,0}
\definecolor{verilogkeywordcolor}{RGB}{49,49,255}
\definecolor{backcolor}{RGB}{237,237,237}
\definecolor{verilogsystemcolor}{RGB}{128,0,255}
\definecolor{verilognumbercolor}{RGB}{255,143,102}
\definecolor{verilogstringcolor}{RGB}{160,160,160}
\definecolor{verilogdefinecolor}{RGB}{128,64,0}
\definecolor{verilogoperatorcolor}{RGB}{0,0,128}
\definecolor{pointcolor}{RGB}{192,0,0} 
\lstdefinestyle{prettyverilog}{
   backgroundcolor=\color{backcolor},
   language           = Verilog,
   commentstyle       = \color{verilogcommentcolor},
   alsoletter         = \$'0123456789\`,
   literate           = *{+}{{\verilogColorOperator{+}}}{1}%
                         {-}{{\verilogColorOperator{-}}}{1}%
                         {@}{{\verilogColorOperator{@}}}{1}%
                         {;}{{\verilogColorOperator{;}}}{1}%
                         {*}{{\verilogColorOperator{*}}}{1}%
                         {?}{{\verilogColorOperator{? }}}{1}%
                         {:}{{\verilogColorOperator{:}}}{1}%
                         {<}{{\verilogColorOperator{<}}}{1}%
                         {>}{{\verilogColorOperator{>}}}{1}%
                         {!}{{\verilogColorOperator{!}}}{1}%
                         {xorsymbol}{{\verilogColorOperator{^}}}{1}%
                         {|}{{\verilogColorOperator{| }}}{1}%
                         {||}{{\verilogColorOperator{|| }}}{1}%
                         {=}{{\verilogColorOperator{= }}}{1}%
                         {==}{{\verilogColorOperator{== }}}{1}%
                         {=>}{{\verilogColorOperator{=> }}}{1}%
                         {[}{{\verilogColorOperator{[}}}{1}%
                         {]}{{\verilogColorOperator{]}}}{1}%
                         {(}{{\verilogColorOperator{(}}}{1}%
                         {)}{{\verilogColorOperator{)}}}{1}%
                         {rightbracket}{{\verilogColorOperator{)}}}{1}%
                         {,}{{\verilogColorOperator{,}}}{1}%
                         {.}{{\verilogColorOperator{.}}}{1}%
                         {~}{{\verilogColorOperator{$\sim$}}}{1}%
                         {\%}{{\verilogColorOperator{\%}}}{1}%
                         {\&}{{\verilogColorOperator{\& }}}{1}%
                         {\&\&}{{\verilogColorOperator{\&\& }}}{1}%
                         {\#}{{\verilogColorOperator{\#}}}{1}%
                         {\ /\ }{{\verilogColorOperator{\ /\ }}}{3}%
                         {\ _}{\ \_}{2}%
                        ,
   morestring         = [s][\color{verilogstringcolor}]{"}{"},%
   identifierstyle    = \color{black},
   vlogdefinestyle    = \color{verilogdefinecolor},
   vlogconstantstyle  = \color{verilognumbercolor},
   vlogsystemstyle    = \color{verilogsystemcolor},
   basicstyle         = \small\fontencoding{T1}\ttfamily,
  columns=fullflexible, 
   keywordstyle       = \bfseries\color{verilogkeywordcolor},
   morekeywords      = {val, when, port, coverage, unique},
   numbers            = left,
   numbersep          = 5pt,
   tabsize            = 2,
   escapeinside       = {/*!}{!*/},
   upquote            = true,
   sensitive          = true,
   showstringspaces   = false, 
   frame              = single, 
   breaklines         = true,
   abovecaptionskip   = 0pt,
   belowcaptionskip   = 0pt, 
   xleftmargin        =0.35cm,
   xrightmargin       =0.15cm,
   captionpos         = b,
   emph               = {Point, Point0, Point1, Point2, Point3, Point4, Point5, Point6, Point7, Point8, Point9},
   emphstyle          =\color{pointcolor},
   emph               = {[2] STVEC,SCOUNTEREN,MSTATUS,MTVEC,ML1_ICACHE_MISS,ML1_DCACHE_MISS,MITLB_MISS,MDTLB_MISS,
                             MLOAD,MSTORE,MEXCEPTION,MEXCEPTION_RET,MBRANCH_JUMP,MCALL,MRET,MMIS_PREDICT,MSB_FULL,
                             MIF_EMPTY,MHPM_COUNTER_17,MHPM_COUNTER_18,MHPM_COUNTER_19,MHPM_COUNTER_20,MHPM_COUNTER_21,
                             MHPM_COUNTER_22,MHPM_COUNTER_23,MHPM_COUNTER_24,MHPM_COUNTER_25,MHPM_COUNTER_26,MHPM_COUNTER_27,
                             MHPM_COUNTER_28,MHPM_COUNTER_29,MHPM_COUNTER_30,MHPM_COUNTER_31}, 
   emphstyle          = {[2]\bfseries\color{verilogkeywordcolor}}
}

\makeatletter

\newcommand\language@verilog{Verilog}
\expandafter\lst@NormedDef\expandafter\languageNormedDefd@verilog%
  \expandafter{\language@verilog}
  
\lst@SaveOutputDef{`'}\quotesngl@verilog
\lst@SaveOutputDef{``}\backtick@verilog
\lst@SaveOutputDef{`\$}\dollar@verilog

\newcommand\getfirstchar@verilog{}
\newcommand\getfirstchar@@verilog{}
\newcommand\firstchar@verilog{}
\def\getfirstchar@verilog#1{\getfirstchar@@verilog#1\relax}
\def\getfirstchar@@verilog#1#2\relax{\def\firstchar@verilog{#1}}

\newcommand\addedToOutput@verilog{}
\lst@AddToHook{Output}{\addedToOutput@verilog}

\newcommand\constantstyle@verilog{}
\lst@Key{vlogconstantstyle}\relax%
   {\def\constantstyle@verilog{#1}}

\newcommand\definestyle@verilog{}
\lst@Key{vlogdefinestyle}\relax%
   {\def\definestyle@verilog{#1}}

\newcommand\systemstyle@verilog{}
\lst@Key{vlogsystemstyle}\relax%
   {\def\systemstyle@verilog{#1}}

\newcount\currentchar@verilog
  
\newcommand\@ddedToOutput@verilog
{%
   \ifnum\lst@mode=\lst@Pmode%
      \expandafter\getfirstchar@verilog\expandafter{\the\lst@token}%
      \expandafter\ifx\firstchar@verilog\backtick@verilog
         \let\lst@thestyle\definestyle@verilog%
      \else
         \expandafter\ifx\firstchar@verilog\dollar@verilog
            \let\lst@thestyle\systemstyle@verilog%
         \else
            \expandafter\ifx\firstchar@verilog\quotesngl@verilog
               \let\lst@thestyle\constantstyle@verilog%
            \else
               \currentchar@verilog=48
               \loop
                  \expandafter\ifnum%
                  \expandafter`\firstchar@verilog=\currentchar@verilog%
                     \let\lst@thestyle\constantstyle@verilog%
                     \let\iterate\relax%
                  \fi
                  \advance\currentchar@verilog by \@ne%
                  \unless\ifnum\currentchar@verilog>57%
               \repeat%
            \fi
         \fi
      \fi
   \fi
}

\lst@AddToHook{PreInit}{%
  \ifx\lst@language\languageNormedDefd@verilog%
    \let\addedToOutput@verilog\@ddedToOutput@verilog%
  \fi
}

\newcommand{\verilogColorOperator}[1]
{%
  \ifnum\lst@mode=\lst@Pmode\relax%
   {\bfseries\textcolor{verilogoperatorcolor}{#1}}%
  \else
    #1%
  \fi
}

\makeatother

\lstdefinestyle{mystyle}{
    commentstyle=\textit,
    keywordstyle=\textbf,
    stringstyle=\color{codepurple},
    basicstyle=\ttfamily,
    breakatwhitespace=false,         
    breaklines=true,      
    frame=single, 
    framexleftmargin=\parindent,
    captionpos=b,                    
    keepspaces=true,                 
    numbers=left,    
    numberstyle=\normalsize,
    stepnumber=1,
    numbersep=5pt,   
    xleftmargin=1.5\parindent,
    showspaces=false,                
    showstringspaces=false,
    showtabs=false,                  
    tabsize=2
}

\usepackage{xurl}

\usepackage{enumitem}

\newcommand{\myname}{\texttt{CreativEval}}

\usepackage[normalem]{ulem}

\usepackage{hyperref}

\begin{document}

\title{\myname{}: Evaluating Creativity of LLM-Based Hardware Code Generation
}

\author{Matthew DeLorenzo, Vasudev Gohil, Jeyavijayan Rajendran\\
Texas A\&M University, USA\\
{\tt \{matthewdelorenzo, gohil.vasudev,  jv.rajendran\}@tamu.edu} 
}

\maketitle

\begin{abstract}
Large Language Models (LLMs) have proved effective and efficient in generating code, leading to their utilization within the hardware design process. Prior works evaluating LLMs' abilities for register transfer level code generation solely focus on functional correctness.
However, the creativity associated with these LLMs, or the ability to generate novel and unique solutions, is a metric not as well understood, in part due to the challenge of quantifying this quality.

To address this research gap, we present \myname{}, a framework for evaluating the creativity of LLMs within the context of generating hardware designs. We quantify four creative sub-components, fluency, flexibility, originality, and elaboration, through various prompting and post-processing techniques. We then evaluate multiple popular LLMs (including GPT models, CodeLlama, and VeriGen) upon this creativity metric, with results indicating GPT-3.5 as the most creative model in generating hardware designs.
\end{abstract}

\begin{IEEEkeywords}
Hardware Design, LLM, Creativity
\end{IEEEkeywords}

\section{Introduction}\label{sec:introduction}
Recent advancements within artificial intelligence, machine learning, and computing performance have resulted in the development of LLMs, which have quickly proven to be a widely applicable and successful solution when applied to a variety of text-based tasks~\cite{LLMs_applications_blog}. After extensive training on large quantities of text data, these transformer-based models~\cite{vaswani2023attention} have demonstrated the ability to not only successfully interpret the contextual nuances of a provided text (or prompt), but also generate effective responses to a near human-like degree~\cite{cai2024large}. This can take the form of summarizing a document, answering and elaborating upon questions, and even generating code. The effectiveness and versatility of LLMs regarding textual understanding have resulted in their adoption within various applications, such as language translation~\cite{kocmi2023large}, customer service chat-bots~\cite{pandya2023automating}, and programming assistants~\cite{LLMs_applications_blog}.

Furthermore, the potential of LLM code generation has recently been explored within the integrated circuit (IC) design process~\cite{zhong2023llm4eda}, such as within the logic design stage. With chip designs continually growing in scale and complexity, efforts to increase the automation of this task through LLMs have been explored. This includes the evaluation of LLMs' ability to generate hardware design codes from English prompts, leading to promising initial results within various frameworks~\cite{thakur2023verigen,Blocklove_2023, liu2024chipnemo,delorenzo2024make}.

With the goal of further optimizing these LLMs to the level of an experienced hardware designer, many research efforts have focused on improving performance within the metric of code functionality. This includes testing various LLM fine-tuning strategies and prompting methods for domain-optimized performance, such as register transfer level (RTL) code generation.

However, another dimension to consider when evaluating the ability of a designer, absent from previous evaluations, is creativity. This term refers to the capacity to think innovatively---the ability to formulate new solutions or connections that are effective and unconventional~\cite{article}. When applied to hardware code generation, this can take the form of writing programs that are not only correct, but also novel, surprising, or valuable when compared to typical design approaches. This quality is essential to understanding the greater potential of LLMs as a tool for deriving new approaches to hardware design challenges, rather than simply a method to accelerate existing design practices. 
With a quantitative method of measuring this concept of creativity within LLM hardware generation, valuable insights could be derived, such as how performance could be further improved, or how LLMs can be best utilized within the hardware design process.

To address this absence within the analysis of LLM-based RTL code generation, we propose a comparative evaluation framework in which the creativity of LLMs can be effectively measured. This assessment is composed of four cognitive subcategories of creativity (fluency, flexibility, originality, and elaboration), which are quantified and evaluated within the context of generating functional Verilog modules. Furthermore, this approach utilizes various prompting structures, generation strategies, and post-processing methods, from which the quality and variations of responses are utilized to generate a metric for creativity. This work presents the following contributions:
\begin{itemize}[leftmargin=*]
\item To the best of our knowledge, we propose the first framework from which a metric for creativity is defined for LLMs within the context of hardware design and code generation.
\item We provide a comparative evaluation between state-of-the-art LLMs upon our creativity metric and its components, with GPT-3.5 achieving the highest result.
\item To enable future research, we will open-source our framework codebase and datasets here: \href{https://github.com/matthewdelorenzo/CreativEval/}{https://github.com/matthewdelorenzo/CreativEval/}
\end{itemize}

\section{Background and Related Work}\label{sec:background}
\subsection{LLMs for Code Generation and Hardware Design}

\begin{figure*}[t]
    \centering
    \includegraphics[width=\textwidth,trim={0 0.2cm 0 0cm},clip]{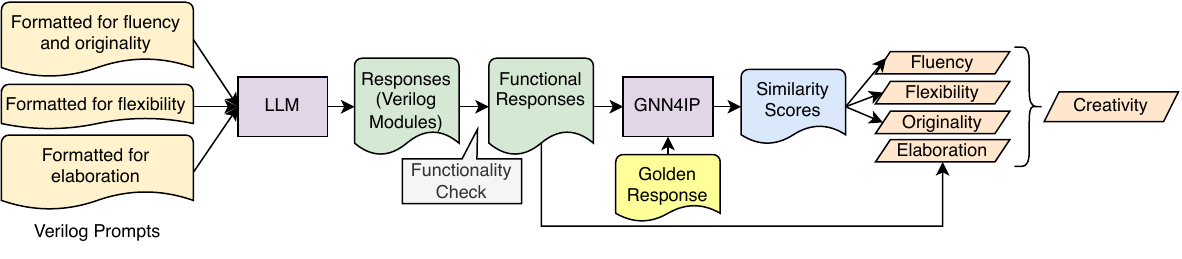}
    \caption{Experimental Framework - calculating creativity of LLMs in Verilog code generation.}
    \label{fig:final_framework_figure}
\end{figure*}

Many state-of-the-art LLMs have demonstrated remarkable success in generating code when provided only with a natural language description, such as GPT-3.5/4 \cite{openai2024gpt4}, BERT \cite{devlin_bert_2019}, and Claude \cite{Anthropic}, revolutionizing the software development process. These models demonstrate promising performance in code functionality, such as GPT-4 generating correct code for 67\% of programming tasks in the HumanEval benchmark in a single response (pass@1)~\cite{luo2023wizardcoder,chen2021evaluating,wang2023codet5}.

Therefore, the applications of LLMs within hardware design through RTL code generation are explored within various studies, such as DAVE~\cite{dave} which utilized GPT-2 for this task. VeriGen~\cite{thakur2023verigen} then demonstrated that fine-tuning smaller models (CodeGen) upon a curated RTL dataset can outperform larger models in RTL tests. VerilogEval~\cite{liu2023verilogeval} presents enhanced LLM hardware generation through supervised fine-tuning, and provides an RTL benchmark for evaluating functionality in RTL generation. ChipNeMo~\cite{liu2024chipnemo} applied fine-tuning upon open-source models (Llama2 7B/13B) for various hardware design tasks. RTLCoder~\cite{liu2024rtlcoder} presents an automated method for expanding the RTL dataset used for fine-tuning, resulting in a 7B-parameter model that outperforms GPT-3.5 on RTL benchmarks. Other works, including RTLLM~\cite{lu2023rtllm} and Chip-Chat~\cite{Blocklove_2023}, explore prompt engineering strategies to enhance the quality and scale of LLM-generated designs. Although there is a plethora of work on LLM-based RTL generation, none of these prior works assess the creative component of LLMs in the hardware design process. We address this shortcoming in this work.

\subsection{Evaluating Creativity}
Prior cognitive science studies~\cite{ALMEIDA200853,10a9e376-29f1-3b1d-ab73-402da965301f,Guilford_1971,torrance1966torrance} have explored methods in which creative thinking can be effectively measured. A widely accepted creativity model~\cite{Guilford_1971} defines four primary cognitive dimensions from which divergent thinking, or the ability to generate creative ideas through exploring multiple possible solutions~\cite{Arefi_2018}, can be measured---fluency, flexibility, originality, and elaboration.

\begin{itemize}[leftmargin=*]
    \item \textbf{Fluency.} The quantity of relevant and separate ideas able to be derived in response to a single given question.
    \item \textbf{Flexibility.} The ability to formulate alternative solutions to a given problem or example across a variety of categories.
    \item \textbf{Originality.} A measure of how unique or novel a given idea is, differing from typical responses or solutions.
    \item \textbf{Elaboration.} The ability to expand upon or refine a given idea. This can include the ability to construct complex solutions utilizing provided, basic concepts.
\end{itemize}
These subcategories have been widely in evaluating human creativity within educational research, including various studies of students~\cite{Handayani_2021,grade_level,arefi2016comparation} as a metric for effective learning. Furthermore, recent works explore the intersection between cognitive science and LLMs~\cite{Shiffrin_Mitchell_2023,stevenson2022putting,Binz_2023}, in which the creativity of LLMs are evaluated within the context of natural language, demonstrating near-human like performance in many cases~\cite{stevenson2022putting}.
In particular,~\cite{zhao2024assessing} utilizes the four creative subcategories to evaluate LLMs across multiple language-based cognitive tasks. However, this framework has not been adapted to LLMs within the context of generating hardware code. To this end, we devise our creativity evaluation framework for LLM-based hardware code generation.

\section{\myname{} Framework}\label{sec:framework}

Given a target LLM, our \myname{} framework, as shown in Fig. \ref{fig:final_framework_figure}, seeks to evaluate the creativity associated with LLMs in hardware code generation. \myname{} evaluates the previously defined subcategories of creativity---fluency, flexibility, originality, and elaboration.
To this end, we query the target LLM with different Verilog-based prompts, and analyze the responses through various methods of post-processing to calculate the desired metrics, as explained below.

\subsection{Fluency}\label{Fluency}
To capture the quantity of relevant and separate ideas in our context, we define fluency as the average number of unique Verilog solutions generated by the target LLM in response to a given prompt. Our prompts contain a brief English description of the module and the module's declaration, as shown in Listing~\ref{listing:listing1}.
Each prompt is provided as input to the LLM, with the response intended to be the completed implementation of the module. As the inference process of LLMs contain variations in the generated responses, we generate $t$ responses for each prompt to estimate the average performance.

\begin{lstlisting}[language=Verilog, label = {listing:listing1}, caption={Fluency/Originality prompt example},style=prettyverilog,float,belowskip=-15pt,aboveskip=20pt,firstnumber=1,linewidth=\linewidth]
//Create a full adder. 
//A full adder adds three bits (including carry-in) and produces a sum and carry-out.

module top_module ( 
    input a, b, cin,
    output cout, sum rightbracket;
\end{lstlisting}

Upon generating all responses, each response is then tested for functionality against the module's associated testbench. If all test cases pass, the module is considered functional. Then, for each prompt, the functional responses (if any) are collected and compared to identify if they are unique implementations.

This is done through GNN4IP~\cite{yasaei2021gnn4ip}, a tool utilized to assess the similarities between circuits. By representing two Verilog modules as a data-flow graph (DFG), GNN4IP generates a similarity score within [-1,1], with larger values indicating a higher similarity. Each correct generated solution from the LLM is input into GNN4IP, and compared to its ideal solution, or ``golden response". Upon the generation of each similarity value for a given prompt, these results are then compared to determine how many unique values are in the response set, indicating the number of distinct solutions.

Given that there are a set of $p$ total prompts in the dataset, the LLM generates $t$ responses for each. After evaluating the functionality of these results, there is then a subset $n$ prompts that contain at least one success (functional module generation). For each of these $n$ prompts, there is a sub-total of the $t$ responses that are functional, defined as $m$. Each of these $m$ functional responses, $r$, are defined as the set $R = \{{r_{1n}, ..., r_{mn}}\}$. The GNN4IP similarity value is then found for each response in $R$, represented as the function $S$. The number of unique similarity values is then determined within the set, and normalized to total $t$ responses. This process is repeated for all $n$ successful prompts and averaged to define the fluency metric $F$ below:

\begin{equation}
F = \frac{1}{n} \sum_{i=1}^{n} \left(\frac{|S(R_i)|}{t}\right)\label{eq_F}
\end{equation}

\subsection{Flexibility}
Flexibility is quantified as the ability of the LLM to generate an alternative implementation of a Verilog module when provided with a solution. The prompts for this metric are constructed for a set of Verilog modules in which a correct solution (the golden response) is included (Listing~\ref{listing:listing2}). The LLM then rewrites the Verilog module, ideally resulting in a functional and unique implementation.

\begin{lstlisting}[language=Verilog, label = {listing:listing2}, caption={Flexibility prompt example},style=prettyverilog,float,belowskip=-15pt,aboveskip=20pt,firstnumber=1,linewidth=\linewidth]
// You are a professional hardware designer that writes correct, fully functional Verilog modules.
// Given the fully implemented example of the Verilog module below:

module true_module( 
    input a, b, cin,
    output cout, sum rightbracket;
    assign sum =  a ^ b ^ cin;
    assign cout = a & b | a & cin | b & cin;
endmodule

// Finish writing a different and unique implementation of the provided true_module in the module below, top_module.
module top_module ( 
    input a, b, cin,
    output cout, sum rightbracket;
\end{lstlisting}

As before, $t$ responses are generated for each of the $p$ total prompts. After all responses are checked for functionality, $n$ prompts have at least one functional response, each with $m$ functional responses. These functional responses are compared directly with the golden response (through GNN4IP) to identify their similarity value. If the similarity value $s$ is lower than a given threshold on the scale [-1,1], the response is considered an alternative solution, shown in Equation \ref{eq_T}. For each successful prompt, the response with minimum similarity is found and evaluated against the threshold. Then, the total amount of $n$ prompts with a response less than the threshold is determined, and normalized by the total prompts $n$. The final flexibility metric $X$ is then defined below:

\begin{equation}
T(s) = \begin{cases} 1 & \text{if } s < 0\\ 0 & \text{if } s \geq 0 \end{cases}\label{eq_T}
\end{equation}

\begin{equation}
X = \frac{1}{n} \sum_{i=1}^{n} \left(T[\min_{}S(R_i)]\right)\label{eq_flex}
\end{equation}

\subsection{Originality}
The originality metric is defined as the variance (uniqueness) of an LLM-generated Verilog module in comparison to a typical, fully functional implementation. This metric is derived from the similarity value (generated through GNN4IP) between successful generations and their golden response.

The originality experiment follows the same prompt structure and procedure as described in \ref{Fluency}. For each prompt, the response with the minimum similarity value is found. Then, the similarity values [-1, 1] are re-normalized to be on scale of [0, 1] with 1 indicating the least similarity (i.e. most original). These results are averaged over all $n$ prompts, with the final originality metric $O$ is described below:

\begin{equation}
O = \frac{1}{n} \sum_{i=1}^{n} \frac{\left(-\min_{} S(R_i)+1\right)}{2}\label{eq_O}
\end{equation}
\subsection{Elaboration}
To measure an LLM's capacity for elaboration, the LLM is provided with multiple smaller Verilog modules in a prompt, and tasked with utilizing them to implement a larger, more complex module. As this metric requires multi-modular designs, a separate set of Verilog modules is utilized in constructing the prompts, as shown in Listing~\ref{listing:listing3}.

\begin{lstlisting}[language=Verilog, label = {listing:listing3}, caption={Elaboration prompt example},style=prettyverilog,float,belowskip=-15pt,aboveskip=20pt,firstnumber=1,linewidth=\linewidth]
// You are given a module add16 that performs a 16-bit addition. 
//Instantiate two of them to create a 32-bit adder. 

module add16 ( input[15:0] a, input[15:0] b, input cin, output[15:0] sum, output cout rightbracket;

module top_module (
    input [31:0] a,
    input [31:0] b,
    output [31:0] sum
rightbracket;
\end{lstlisting}

\begin{table*}[htbp]
    \caption{Comparison of different LLMs in terms of creativity and its subcategories}
    \centering
    \resizebox{\textwidth}{!}{%
    \begin{tabular}{lcccccc} 
        \toprule
        LLM & Functionality & Fluency & Flexibility & Originality & Elaboration & Creativity\\
        \hline
        \hline
        CodeLlama-7B \cite{codellama/CodeLlama-7b-hf}& 0.2417 & 0.1483 & 0.0000 & 0.2926 & 0.2222 & 0.1658\\ 
        \hline
        CodeLlama-13B \cite{codellama/CodeLlama-13b-hf} & 0.3167 & 0.1611 & 0.0260 & \textbf{0.3021} & \textbf{0.3333} & 0.2056 \\
        \hline
        VeriGen-6B \cite{shailja/fine-tuned-codegen-6B-Verilog} & 0.3667 & 0.1244 & 0.1000 & 0.2527 & \textbf{0.3333} & 0.2026\\
        \hline
        VeriGen-16B \cite{shailja/fine-tuned-codegen-16B-Verilog}& 0.3250 & 0.1189 & 0.0556 & 0.2771 & \textbf{0.3333} & 0.1962\\
        \hline
        GPT-3.5 \cite{GPT-3.5}& 0.3083 & 0.1343 & \textbf{0.1600 }& 0.2526 & \textbf{0.3333} & \textbf{0.2201}\\
        \hline
        GPT-4 \cite{GPT-4} & \textbf{0.3750} & \textbf{0.1644} & 0.0795& 0.2657 & \textbf{0.3333} & 0.2107\\
        \bottomrule
    \end{tabular}
    }
    \label{tab1}
\end{table*}
Multiple LLM responses are generated for each module, which are all then checked for functionality. For all given functional solutions, the responses are checked to see if the solution utilizes the smaller modules (as opposed to a single modular solution). If any of the responses for a given prompt are both functional and utilize the smaller modules, it is considered a positive instance of elaboration. Given $p$ total Verilog prompts, of which $n$ have at least one response that demonstrates elaboration, the metric is specified as:
\begin{equation}
 E =\left(\frac{n}{p}\right)\label{eq_E}
\end{equation}
\subsection{Creativity: Putting It All Together}
Given each of the subcategories associated with creativity defined above, the metrics are then combined to define the overall creativity of a given LLM in Verilog hardware design.
\begin{equation}
 C =(0.25)F +(0.25)X + (0.25)O + (0.25)E\label{eq_C}
\end{equation}

\section{Experimental Evaluation}
\subsection{Experimental Setup}\label{exp_setup}
We evaluate multiple LLMs using the \myname{} framework, including CodeLlama 7B \cite{codellama/CodeLlama-7b-hf} and 13B \cite{codellama/CodeLlama-13b-hf} parameter models, VeriGen 6B \cite{shailja/fine-tuned-codegen-6B-Verilog} and 16B \cite{shailja/fine-tuned-codegen-16B-Verilog} (16B model loaded in 8-bit quantization due to memory constraints), GPT-3.5 \cite{GPT-3.5}, and GPT-4 \cite{GPT-4}. The inference process of the VeriGen and CodeLlama models was performed locally on an NVIDIA A100 GPU with 80 GB RAM, while GPT-3.5/4 were queried through the OpenAI Python API.  All scripts are written in Python 3.10, with Icarus Verilog 10.3  as the simulator for evaluating functionality checks.
The open-source GNN4IP repository was adapted to this framework to generate the similarity scores. The prompt dataset utilized for functionality, fluency, and originality consists of 111 single-module HDLBits~\cite{HDLBITS} prompts sourced through AutoChip~\cite{thakur2023autochip}, each containing a correctly implemented solution and testbench. The smaller prompt set used for elaboration contains 9 separate multi-module prompts from the same source. The base functionality metric (pass@10) is measured on all 120 prompts.

When generating LLM responses in all experiments, the LLMs were all set to the following inference hyperparameters: temperature=0.3; max\_tokens=1024; top\_k=10; top\_p=0.95. All responses were trimmed to the first generated instance of ``endmodule" for effective functionality evaluation.

\subsection{Results}
Table~\ref{tab1} summarizes the results for all LLMs for all subcategories of creativity. In evaluating \textbf{fluency}, GPT-4 had the highest quantity of separate and correct Verilog solutions to a module (with respect to the modules that have at least one correct solution), with CodeLlama-13B achieving similar results. The VeriGen models comparatively struggled in this metric, partly due to repeated generations of similar implementations instead of different implementations.

Regarding \textbf{flexibility}, GPT-3.5 had the highest rate of generating alternative solutions to provided modules across most models. The models that struggled (e.g., CodeLlama) produced results that were often direct copies of the provided module, indicating the ability to understand the prompt's natural language description as an important factor that determined flexibility.

As for \textbf{originality}, the GPT models had slightly worse performance than the others, with CodeLlama performing best. This means that the successful solutions provided with the GPT models were, on average, closer to the ideal solution. This could be due to its large size and training dataset, resulting in a more direct retrieval of existing solutions or coding practices.

\textbf{Elaboration} was largely similar for all modules, as the HDLBits dataset for this metric is comparatively small (9 modules). The models primarily excelled in correctly connecting the input and output parameters between separate modules, while struggling to generate the larger module solution.

Overall, the GPT models were the most \textbf{creative}, with GPT-3.5 as the best, and CodeLlama-7B was the least creative. Creativity is shown to slightly drop for the larger model sizes of GPT and VeriGen.

\section{Conclusion}
Recent studies on LLMs regarding their applications to hardware design have effectively demonstrated their potential, applying many optimization strategies to increase the performance in terms of functional correctness. However, these studies do not investigate the creativity associated with LLMs in their ability to generate solutions, largely due to the lack of an effective metric. Within this work, we propose \myname{}, a framework to evaluate the creativity of LLMs in generating hardware code. By evaluating multiple popular LLMs within this framework, we perform a comparative analysis, concluding that GPT-3.5 had the greatest creativity. Future research in this direction can further evaluate more LLMs and on larger prompt sets.
\section*{Acknowledgment}
\noindent The authors acknowledge the support from the Purdue Center for Secure Microelectronics Ecosystem – CSME\#210205. This work was also partially supported by the National Science Foundation (NSF CNS--1822848 and NSF DGE--2039610).

\bibliographystyle{IEEEtran}
\bibliography{main.bib}

\end{document}